\documentclass[10pt,twocolumn,letterpaper]{article}

\usepackage{iccv}
\usepackage{times}
\usepackage{epsfig}
\usepackage{graphicx}
\usepackage{amsmath}
\usepackage{amssymb}

\usepackage{slashbox}
\usepackage{booktabs}
\usepackage{multirow}
\usepackage{scrextend}
\usepackage[symbol]{footmisc}
\usepackage{tabularx}

\usepackage[breaklinks=true,bookmarks=false]{hyperref}

\iccvfinalcopy 

\ificcvfinal\fi

\begin{document}

\title{Pose Guided Attention for Multi-label Fashion Image Classification \\[3ex] 
  {\normalfont\large 
    Beatriz Quintino Ferreira\textsuperscript{1}, Jo\~{a}o P. Costeira\textsuperscript{1}, Ricardo G. Sousa\textsuperscript{2}, Liang-Yan Gui\textsuperscript{3}, Jo\~{a}o P. Gomes\textsuperscript{1}%
  \thanks{This work was partially funded by FCT via grant [PD/BD/114430/2016] and project [UID/EEA/50009/2019]. We also thank the guidance and support provided by Professor Jos\'{e} F. Moura.}}
  \vspace{-1em}}




\author{
\textsuperscript{1} ISR-IST, Universidade Lisboa, 
\textsuperscript{2} Farfetch, \textsuperscript{3} Carnegie Mellon University \\ {\tt\small \{beatrizquintino, jpc, jpg\}@isr.ist.utl.pt, ricardo.sousa@farfetch.com, lgui@andrew.cmu.edu}}


\maketitle

\ificcvfinal\thispagestyle{empty}\fi

\begin{abstract}
We propose a compact framework with guided attention for multi-label classification in the fashion domain. Our visual semantic attention model (VSAM) is supervised by automatic pose extraction creating a discriminative feature space. VSAM outperforms the state of the art for an in-house dataset and performs on pair with previous works on the DeepFashion dataset, even without using any landmark annotations.
Additionally, we show that our semantic attention module brings robustness to large quantities of wrong annotations and provides more interpretable results.
\end{abstract}
\vspace{-1.4em}
\section{Introduction}
Image classification is a fundamental Computer Vision task, widely applied in the fashion industry to generate rich product descriptions and automate product tagging. This automation is pivotal when dealing with extremely large collections, which naturally arise in most e-commerce platforms. With the advent of Deep Convolutional Neural Networks (CNN's), combined with the availability of massive amounts of data, the ability to extract categories and attributes from visual data gained extreme relevance. However, the  performance gains on the computing side usually come with a high cost from human-intensive tasks to generate high quality training data, a key (and expensive) issue in the fashion industry.

Fashion attributes are often associated to specific locations (e.g short sleeve, neckline). This knowledge is either disregarded in purely data-driven approaches~\cite{QuintinoFerreira2018, Zhu2017} or requires the mentioned expensive annotation processes~\cite{Wang2018AttFashionGrammar, Liu2019}. While the former lacks  interpretability and robustness to image artifacts, the latter is impractical at very large scale.

\begin{figure}[tb]
\centering
\includegraphics[width=0.45\textwidth]{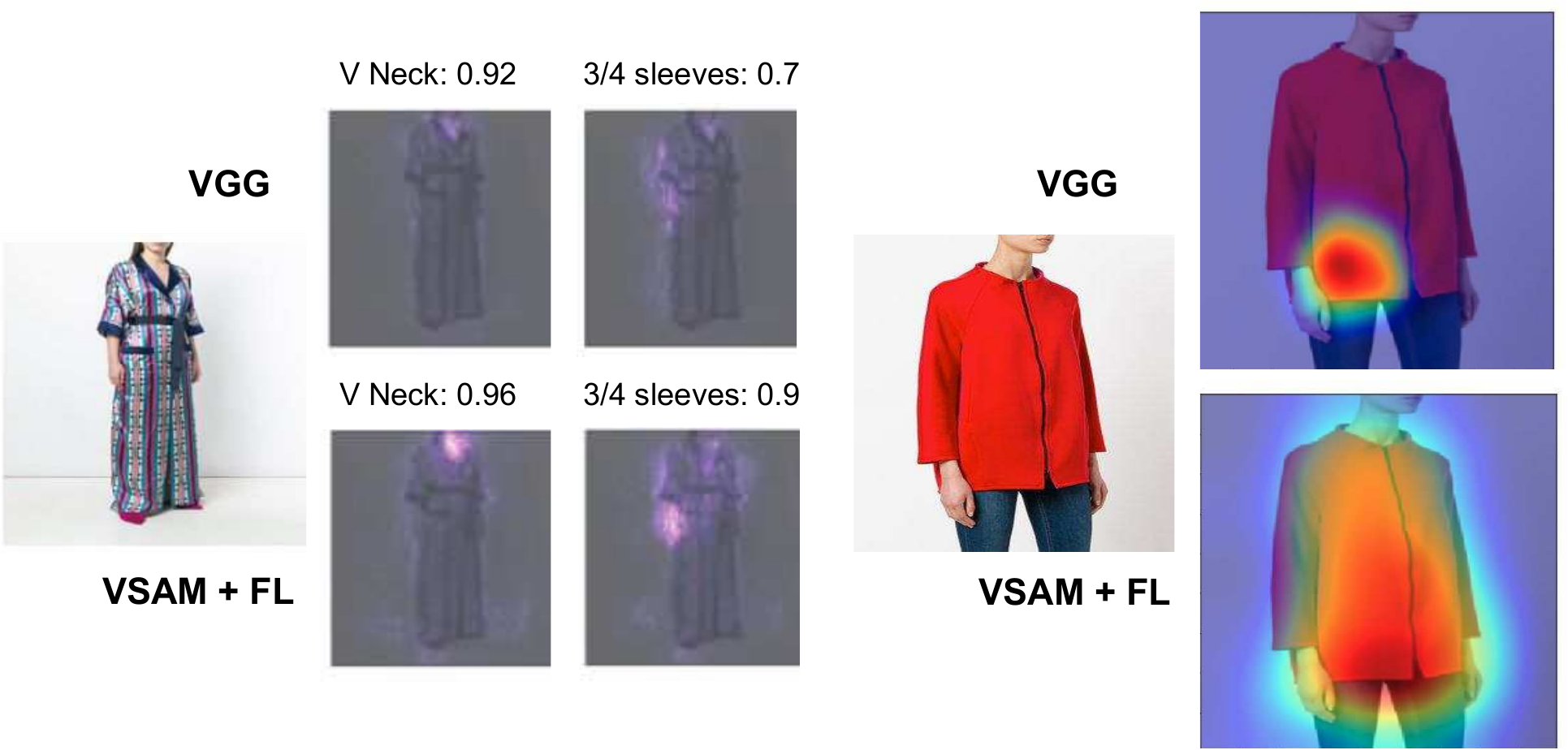}
\caption{The effect of the pose-guided attention: left - saliency maps highlight neck and arm regions for V-neck and 3/4 sleeves attributes, respectively; right - CAMs highlight the relevant region for Jackets category. }
\vspace{-1em}
\label{fig:saliency_CAMs_maps}
\end{figure}

In this work we propose a CNN model that jointly learns to predict fashion categories (multi-class problem) and attributes (multi-label problem) by focusing on the relevant image regions through a guided attention mechanism. Our approach is premised on the hypothesis that classification tasks benefit if the model identifies salient image regions amplifying their influence, while suppressing irrelevant and potentially confusing information in other regions (see Fig.~\ref{fig:saliency_CAMs_maps}).

We exploit the relation between attribute localization and visual appearance by embedding a semantic attention module guided by body pose. Building on an off-the-shelf architecture (VGG) and adding much less complexity, our model achieves similar performance to state-of-the-art models that use extra (costly) annotations and surpasses by a large margin those who do not. Specifically, our method outperforms previous approaches~\cite{QuintinoFerreira2018} for their "in-house" dataset and is on pair with the state of the art for the DeepFashion dataset~\cite{LiuDeepFashion2016}.
We note that the supervision of the semantic attention comes at a very low cost, as there are readily available pose detectors~\cite{Cao2017openpose} providing high accuracy and nearly real-time detections. 

Finally, we demonstrate that the semantic attention module increases the robustness to large quantities of wrong/missing annotations, that constitutes a prevailing issue in very large datasets.


This supervised attention guides the model to learn a feature space that is more suitable and robust (in terms of inter-class and inter-label confusion) for this specific problem of fashion items classification. 
In contrast to previous works~\cite{Zhu2017,Wang2018AttFashionGrammar, Liu2019}, we learn our attention in a supervised manner that encodes the underlying context and semantics of the fashion image classification problem. 


\vspace{-0.5em}
\section{Related work}

Currently, CNNs are known to achieve leading performance for the multi-class and multi-label problems, and numerous works have recently addressed these problems in the context of fashion analysis, see for example~\cite{LiuDeepFashion2016, Wang2018AttFashionGrammar,QuintinoFerreira2018,Gutierrez2019,Liu2019}.   
The work in~\cite{Gutierrez2019} proposes binary models, where each model predicts a fashion attribute, and introduces product type classifiers to mitigate outlier impact, as a pre-validation step before deciding on an attribute. However, this pipeline of models is neither efficient nor scalable to large datasets with numerous categories and attributes. On the contrary, in our previous work~\cite{QuintinoFerreira2018}, we contributed with a unified model that jointly outputs category, subcategory and attributes predictions leveraging the hierarchical category tree structure to explore label relations. This model improved performance over a pipeline of state-of-the-art models performing the three classification tasks individually and independently.

Attention mechanisms have become very popular in deep learning to help focusing the learning process on the parts of the inputs that are relevant to the task, contributing not only to significant performance improvements but also to
model interpretability. Accordingly, a model
with the right attention should concentrate on
the image regions that are discriminative to the
classification (e.g., to classify a long sleeve, the
model is supposed to focus on the regions
that contain the sleeve)~\cite{Zhu2017,Wang2018AttFashionGrammar}.
Specifically in the fashion domain,~\cite{Liu2019} proposes an attentive fashion network with up-sampled feature maps for landmark localization, category classification and attribute prediction that is shown to outperform previous models on the DeepFashion dataset~\cite{LiuDeepFashion2016}. Contrary to~\cite{Wang2018AttFashionGrammar}, the method in~\cite{Liu2019} generates higher-resolution maps and combines separate attention branches in a unified branch, acting as a soft constraint to the model. However, the inclusion of these branches causes a drastic increase in complexity and parameters. 

The previous examples implement self-learned/ unsupervised attention modules. Nevertheless, when there is \emph{a priori} knowledge, this knowledge can be used to learn the attention in a supervised manner.
In~\cite{Linsley2019WhatWhereAttend} a supervised attention module is also added to off-the-shelf classification CNNs to increase recognition performance.
However, as opposed to~\cite{Linsley2019WhatWhereAttend}, where the ground-truth attention heatmaps are crowd-sourced (human derived click masks), we learn our attention heatmaps in an automatic manner resorting to the pose extracted by a pose detector.


\section{Semantic attention model for fashion images classification}
Our task is to predict a category $C$ and an attribute vector $A$ for each image. Category classification is posed as a multi-class problem, thus $C$ satisfies $0 \leq C \leq n_c -1$, where $n_c$ is the total number of classes. At the attributes level we solve a multi-label problem with $A = (a_1, ... , a_{n_a}), a_i \in \{0,1\}$, where $n_a$ is the total number of attributes and $a_i = 1$ indicates that the image has the i-th attribute. 

\begin{figure}[htb]
\centering
\includegraphics[width=0.4\textwidth]{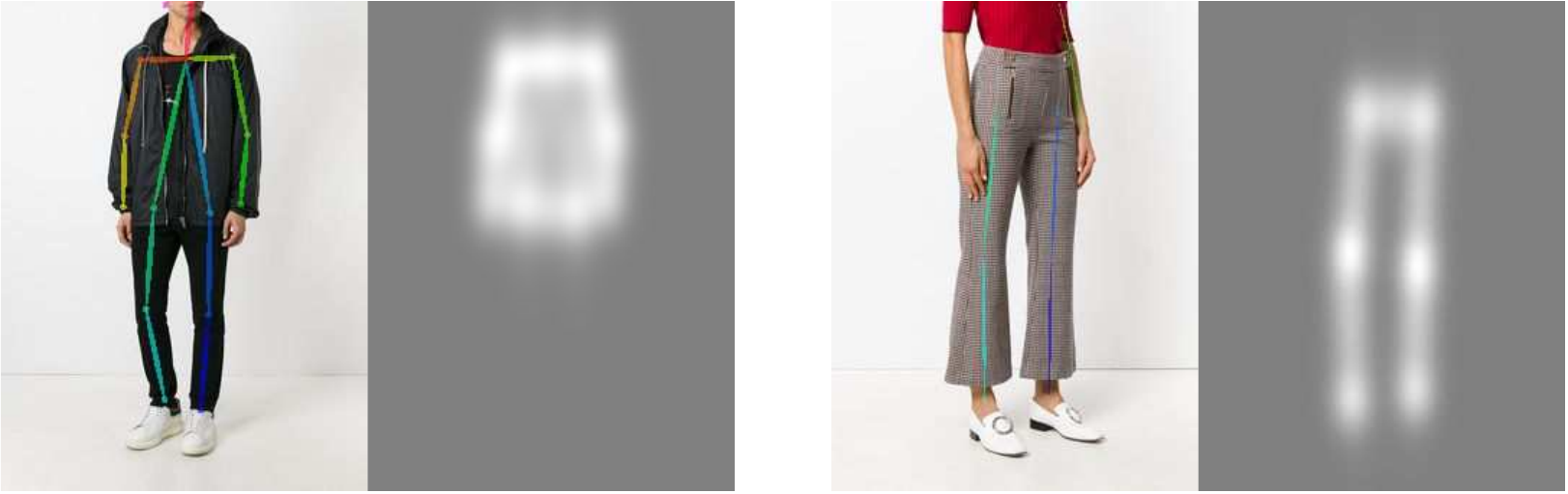}
\caption{Examples of pose detections and respective heatmaps, used during training, with the relevant joints to classify the clothing items in the images.}
\label{fig:pose_heatmap}
\end{figure}
\vspace{-0.3em}

\begin{figure*}[htb]
\label{fig:model_attention}
\centering
\includegraphics[width=0.7\textwidth]{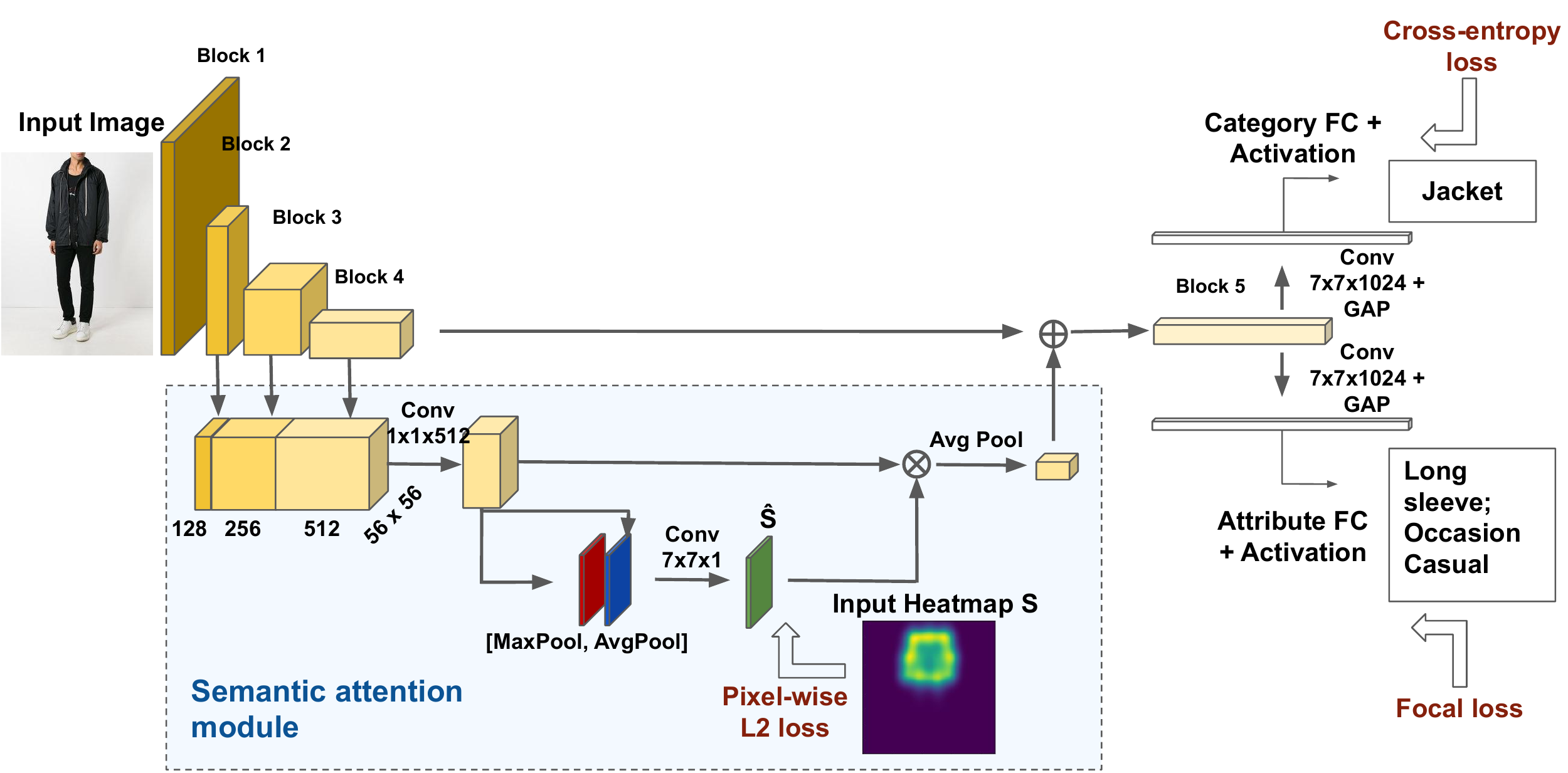}
\caption{Proposed VSAM architecture with the semantic attention mechanism regularization.}
\end{figure*}

Current frameworks incorporate attention modules to augment off-the-shelf classification Neural Nets, both in an unsupervised manner~\cite{Zhu2017,Woo2018CBAM} or with supervision~\cite{Linsley2019WhatWhereAttend}.
In our case, this mechanism acts on a feature combination scheme used in~\cite{Zhang2017TrafficDensity} and is supervised by the pose of the human model wearing the clothing item. The pose is extracted using the off-the-shelf pose detector \textit{OpenPose}~\cite{Cao2017openpose}. 
Given the joint positions provided by the pose detector, we create a heatmap that localizes the clothing item in the image. In particular, these ground-truth heatmaps, used in training, are generated by placing 2D Gaussian filters on the joint locations and connections as shown in Fig.~\ref{fig:pose_heatmap}. 

The proposed VSAM model architecture is depicted in Fig.~\ref{fig:model_attention}.
Specifically, we regularize the backbone network (VGG-16~\cite{Simonyam2015VGG}) with ground-truth heatmaps with the relevant joints obtained by \emph{OpenPose} (see Fig.~\ref{fig:pose_heatmap}). 
This regularization is performed on the feature combination from convolutional blocks 2, 3 and 4 that mix different image resolutions, which are then regularized by the ground-truth heatmap. We also use a spatial attention module (a concatenation of max and average pooling across the channel axis),  similarly to~\cite{Woo2018CBAM}, to highlight informative regions before the regularization. The convolutional layer after the spatial attention module has a sigmoid activation so that the feature values are in the same interval as the heatmaps values (i.e. $ [0,1]$).

The model is trained by minimizing the cross entropy loss at the category level and the focal loss~\cite{LinICCV2017FocalLoss} at the attributes level. The latter has been shown to be effective in preventing excessive negative examples from hindering model training due to the large imbalance of negative/positive examples that arise from annotation sparsity. Finally, for the attention regularizer we use the pixel-wise L2-norm difference between the estimated ($\hat{S}$) and the ground-truth ($S$) heatmaps $\sum_u^{w} \sum_v^h \| \hat{S}(u,v) - S(u,v)\|_2$ .



Our model has two hyperparameters, one weights the contribution of the attention branch, and the other, the heatmap fidelity parameter, multiplies the loss term of the pixel-wise difference between $\hat{S}$ and $S$.  

\section{Experiments and Results Analysis}

\subsection{Experimental Setting}

\paragraph{Datasets:}We use the same (in-house) high-quality dataset from~\cite{QuintinoFerreira2018}, which has approximately 245k images with $n_c = 17$ and $n_a = 53$ and an average number of 1.2 attribute annotations per product. Despite manually curated, some inconsistencies and missing labels naturally arise in this context, leading to a weakly annotated dataset.
We train the proposed model (in Fig.~\ref{fig:model_attention}) on front model images to which we apply OpenPose, and we used a 75\% / 25\% train/test split ratio.

To provide results on a common benchmark, we also experiment with the  widely used DeepFashion dataset~\cite{LiuDeepFashion2016}, containing approximately 289k images, $n_c=50$ and $n_a=1000$. 
Nonetheless, note that, unlike the methods in~\cite{LiuDeepFashion2016, Wang2018AttFashionGrammar, Liu2019}, ours does not use any kind of landmark annotations. 
\vspace{-1.5em}
\paragraph{Compared methods:}
For the in-house dataset, we compare our model with the state-of-the-art model from~\cite{QuintinoFerreira2018}, which had a ResNet-50 as backbone CNN.
We also compare our model with~\cite{LiuDeepFashion2016} that introduced DeepFashion dataset and with the current best-performing model from~\cite{Liu2019}. 
\vspace{-1.5em}
\paragraph{Metrics:}
We focus on the multi-label problem (attributes level) for both datasets. For the in-house dataset we report precision, recall and F1-score at top-k (P@k, R@k, F1@k, where k is the number of ground-truth labels of each product), as well as average precision (AP). 
For the DeepFashion dataset we follow the same evaluation settings from~\cite{LiuDeepFashion2016,Liu2019} and report top-k recall for attribute prediction.
For all these metrics, the larger value, the better the performance. 

All models were run in a single-shot manner (end-to-end) under equal conditions, i.e., for 40 epochs with the same optimizer (Adam with initial learning rate $10^{-4}$ and decay $10^{-5}$), batch size 64 of images resized to $224 \times 224$, and augmentation ratio 0.33. When applied, we used the focal loss standard parameters ($\gamma = 2$, $\alpha = 0.6$).

\subsection{Quantitative results}
\paragraph{In-house dataset:} For categories and considering precision, recall and F1-score, the performance of the proposed method is on pair with the performance of the model from~\cite{QuintinoFerreira2018} that was already very high. 

\begin{table}[!t]
\small
\centering
\resizebox{\columnwidth}{!}{%
\begin{tabular}{ l | c c c c}
\toprule
\backslashbox{Method}{Metric} & \textbf{P$@$k}  & \textbf{R$@$k} & \textbf{F1$@$k} & \textbf{AP} \\ \midrule
Model from~\cite{QuintinoFerreira2018}  & 73.02 & 75.26 & 73.33 & 69.17 \\
\textbf{VSAM + FL} & \textbf{80.70} & \textbf{81.56} & \textbf{80.63} & \textbf{75.44} \\
\toprule
\end{tabular}
}
\caption{Experimental results for multi-label classification.}
\label{tab:results_att_attention_all}
\end{table}

\begin{table*}
\centering
\small
\begin{tabular}{ l | c c c c c c c c}
\toprule
\multirow{2}{*}{\backslashbox{Method\kern-0.7em}{\kern-0.7emMetric}} & \multicolumn{2}{c}{Texture} & \multicolumn{2}{c}{Fabric}  & \multicolumn{2}{c}{Shape}  & \multicolumn{2}{c}{Part} \\ \cmidrule{2-9} & \textbf{top-3} & \textbf{top-5} &\textbf{top-3} & \textbf{top-5} &\textbf{top-3} & \textbf{top-5} &\textbf{top-3} & \textbf{top-5}\\ \midrule
Model from~\cite{QuintinoFerreira2018} & 44.39 & 53.91 &  31.82 & 41.70 & 39.88 & 50.51 & 31.11 & 40.76   \\ 
FashionNet~\cite{LiuDeepFashion2016}\footnote[2] & 37.46 & 49.52 & 39.30 & 49.84 & 39.47 & 48.59 & 44.13 & 54.02\\
Liu et al.~\cite{Liu2019}\footnote[2]  & 56.30 & 65.82 & 43.05 & 53.64 & 58.75 & 67.80 & 46.47 & 57.39 \\
VSAM + FL & 56.28  & 65.45 & 41.73 & 52.01 & 55.69 & 65.40 & 43.20 & 53.95\\
\toprule
\end{tabular}
\caption[Experimental results for attribute classification for the DeepFashion dataset.]{Experimental results for attribute classification for the DeepFashion dataset.\footnote[3]}
\label{tab:results_cat_attention_deepfashion}
\end{table*}


More importantly, Table~\ref{tab:results_att_attention_all} reports the multi-label performance results at the attributes level for the same dataset. As shown, our VSAM-FL semantic attention and focal loss model outperforms the baseline by a large margin for all metrics. 

\begin{figure}[htb]
\centering
\includegraphics[width=0.3\textwidth]{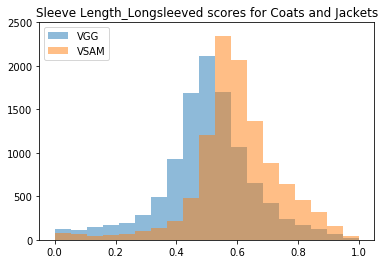}
\caption{Histograms of the scores for 75\% missing label attribute \emph{Longsleeved} for Coats and Jackets categories}
\label{fig:histogram_scores}
\end{figure}

\vspace{-1.5em}
\paragraph{Robustness to wrong annotations:}
In the dataset from~\cite{QuintinoFerreira2018}, the \emph{Longsleeved} attribute is not commonly used for Coats and Jackets retrieval, thus these labels are set to zero ($a_{ GT\_\emph{Longsleeved}} = 0)$ to save annotation effort. 
To study the impact of positive annotations, we modify 25\% of the ground-truth annotations of this attribute to the correct value 1. In Fig.~\ref{fig:histogram_scores} we observe a larger shift of the score's mass of the proposed method towards higher values compared with the VGG (our model without semantic attention). For example, for a decision threshold of $0.5$ the recall of the model with semantic attention is 84\% whereas for the VGG is 52\% (for 0.4 the recall for the proposed model is 13\% higher than for the VGG, and for 0.6 is 25\% higher). 
This suggests that the semantic attention module can have a significant impact in model robustness to wrong annotations which are recurrent in this industry.
\vspace{-0.3em}
\vspace{-1.5em}



\paragraph{DeepFashion:} Table~\ref{tab:results_cat_attention_deepfashion} reports the results on the DeepFashion dataset. Despite the simpler machinery (approx. 25M parameters vs. 76M of~\cite{LiuDeepFashion2016}) and less information, as it does not use landmarks, our model achieves nearly the same results as the top performing from the state of the art.  

\footnotetext{$^\dagger$ method uses landmarks} 
\footnotetext{$^\ddagger$ we used the code kindly made available by~\cite{Liu2019}. We did not include results for Style attributes because were not able to reproduce them.} 



\subsection{Qualitative results}

The attention regularized model hallucinated good attention heatmaps (correct pose) for test images (feature maps appear to focus on relevant locations), and visualization techniques as saliency maps or CAMs~\cite{Selvaraju2016GradCAM} highlight more meaningful regions than for the VGG, as shown in Fig.~\ref{fig:saliency_CAMs_maps} and in the appendix.

\section{Conclusions}

Creating complete and consistent datasets seems an unrealistic task, since fully labeling fashion attributes and, specially, spatial landmarks with human annotators would require a tremendous amount of effort and be extremely expensive. As a consequence, taking advantage of \emph{a priori} cues to look for these details in the images appears as a promising strategy. 
In this work we introduced a fashion classification model, VSAM, whose attention is guided by the pose of the human wearing the clothing items. In spite of its much lower complexity, VSAM outperformed, by a large margin, a previous model in the challenging multi-label scenario for a fashion e-commerce platform dataset (without landmark annotations), and performed on pair with the state-of-the-art methods for the DeepFashion dataset that benefit from landmark annotations. Furthermore, the proposed model was robust to wrong annotations and provided more meaningful visualizations and  interpretability. The encouraging results suggest that additional gains are attainable by learning attribute specific attention maps.


{\small
\bibliographystyle{ieee}
\bibliography{egbib}
}

\onecolumn

\section{Appendix}

\begin{figure*}[!htb] 
\centering
\includegraphics[width=0.8\textwidth]{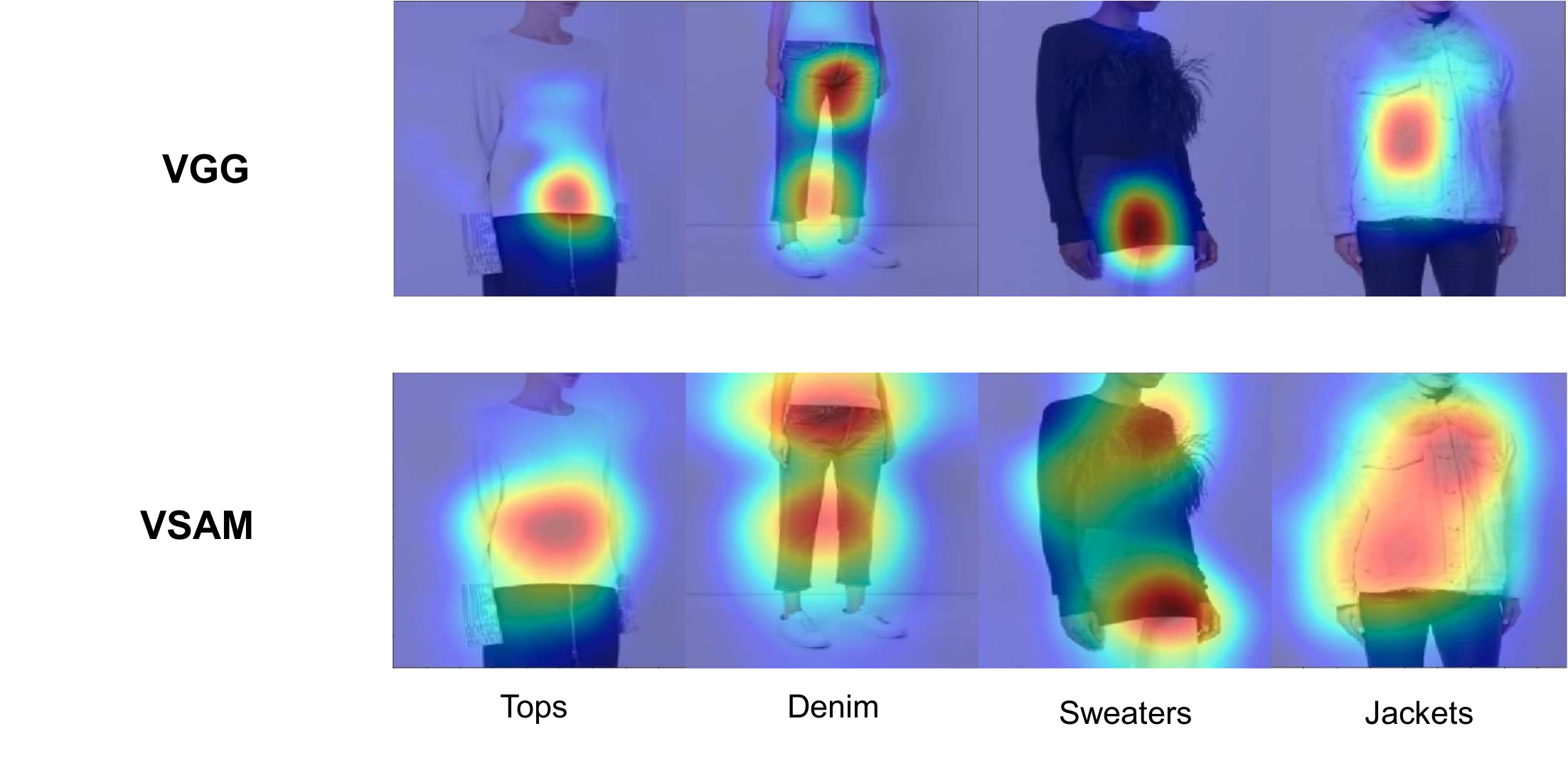}
\caption{Examples of CAMs for different categories for the VGG (without the semantic attention module) and for the proposed model with semantic attention - VSAM.}
\label{fig:cams_appendix}
\end{figure*}

\begin{figure*}[h]
\centering
\includegraphics[width=0.75\textwidth]{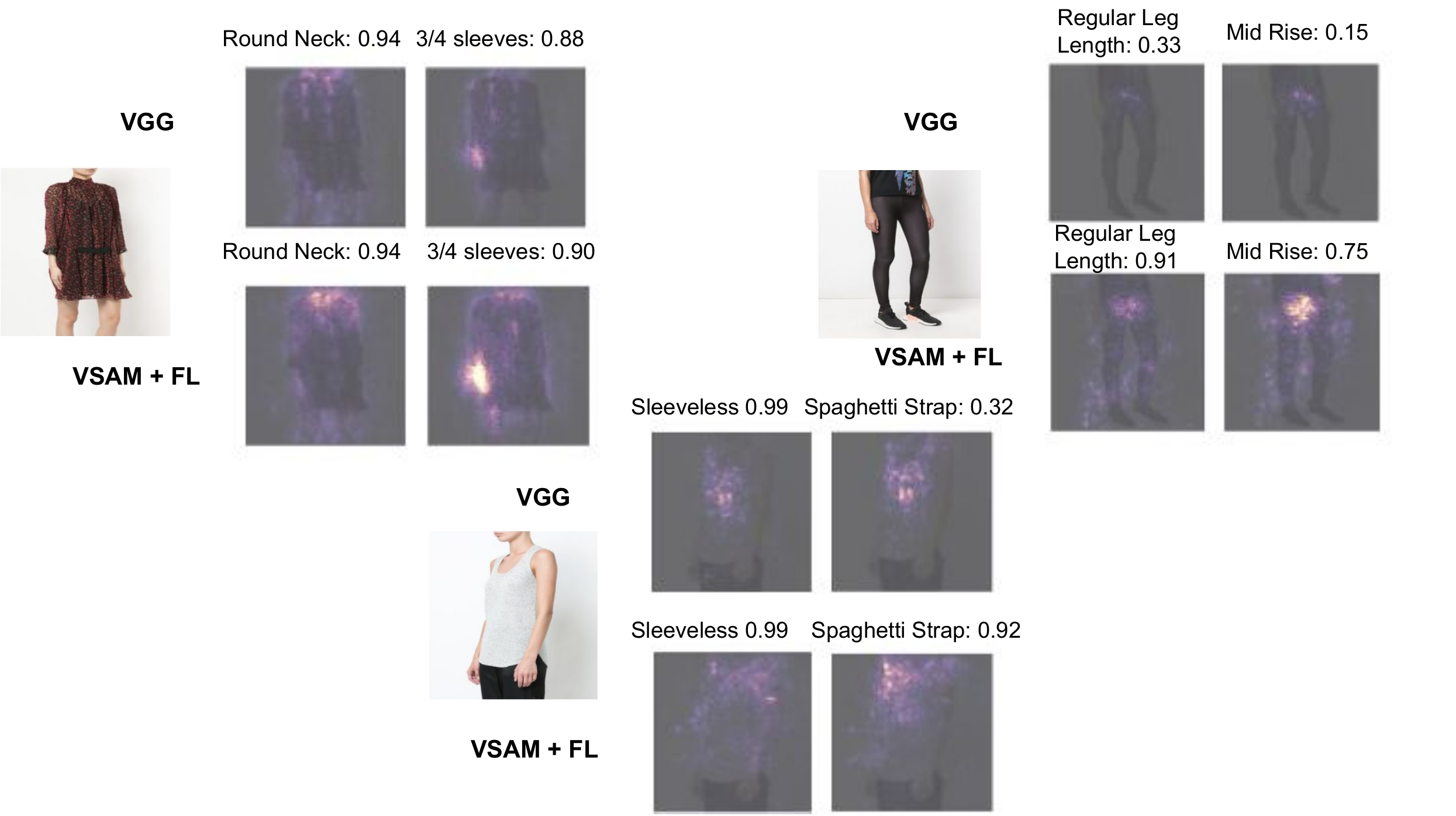}
\caption{Examples of saliency maps for different attributes for the VGG (without the semantic attention module) and for the proposed model with semantic attention - VSAM.}
\label{fig:hallucinated_skeletons}
\end{figure*}

\begin{figure*}[h]
\centering
\includegraphics[width=0.5\textwidth]{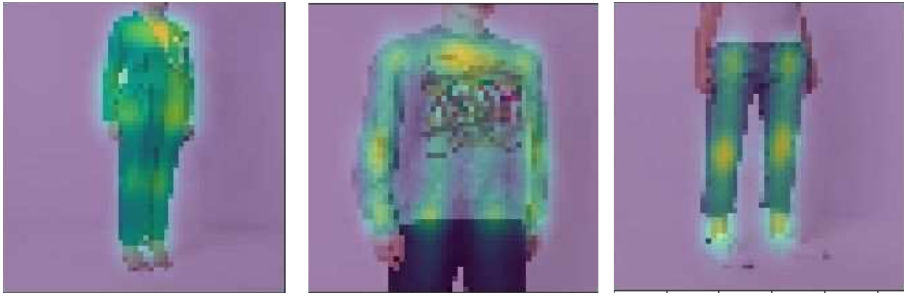}
\caption{Examples of hallucinated poses by VSAM for test images, at the estimated pose heatmap ($\hat{S}$).}
\label{fig:hallucinated_skeletons}
\end{figure*}

\end{document}